\documentclass[conference]{IEEEtran}

\usepackage{amsmath}
\usepackage{booktabs}
\usepackage{tikz}
\usepackage{url}
\usetikzlibrary{arrows.meta,positioning,fit}

\title{Extending FAIR Principles with Observability for Trustworthy Federated Learning}
\author{
    \IEEEauthorblockN{Hangli Ge\IEEEauthorrefmark{1}, 
    Noboru Koshizuka\IEEEauthorrefmark{1}}
    \IEEEauthorblockA{
\IEEEauthorrefmark{1}Interfaculty Initiative in Information Studies; The University of Tokyo}
\IEEEauthorrefmark{1}\{hangli.ge, noboru\}@koshizuka-lab.org;
}
\makeatletter
\def\blfootnote{\xdef\@thefnmark{}\@footnotetext}
\makeatother
\begin{document}
\maketitle

\begin{abstract}
Federated learning (FL) keeps data at its source, but locality alone
does not make a federation trustworthy. Without sufficient observability, an aggregator cannot determine whether the client service was healthy, the local data pipeline was trustworthy, or contractual constraints were properly monitored during training. We take the position that trustworthy FL needs an evidence plane in addition to its model-update process.  Each participant continuously summarizes local infrastructure, data into minimal, signed trust evidence associated with each FL round, while raw data, logs remain local. A local TinyLLM may assist in interpreting observability signals, deriving contextual evidence, and providing natural-language explanations to operators. We instantiate this vision through FAIR-O, an observability extension of FAIR principle. Rather than benchmarking FL accuracy, a three-level trust model has been proposed and a two-participant Python prototype demonstrates evidence generation and trust assessment. We further discuss future integration of FAIR-O Trust Evidence into participant selection, evidence weighting, and federated aggregation.
\end{abstract}

\begin{IEEEkeywords}
Federated Learning, FAIR Principle, Observability, Trust Evidence, TinyLLM
\end{IEEEkeywords}

\section{The Position}
\label{sec:position}
Federated learning (FL) enables multiple clients to train a shared model without centralizing their raw data~\cite{McMahan2017,Kairouz2021}. This separation is especially attractive in industrial systems, where sensor traces, maintenance records, production parameters, and other operational data are distributed across organizations and subject to confidentiality and sovereignty constraints~\cite{Nguyen2021IIoT}. However, secure aggregation can hide individual updates~\cite{Bonawitz2017}, and robust aggregation can tolerate classes of malicious behavior~\cite{Blanchard2017}, yet neither tells an operator whether a particular local training pipeline was observable and trustworthy when its update was produced.

Protecting raw training data alone is therefore insufficient to establish trust in real-world industrial FL. Participants operate over heterogeneous and continuously changing infrastructures, data pipelines, software environments, and governance regimes. A participant may remain technically connected to an FL federation while experiencing degraded computational or network resources, stale or incomplete data, schema changes, provenance gaps, or violations of agreed data-use policies. These conditions may directly affect the reliability and interpretation of a local model update without necessarily appearing as an explicit attack. Consequently, trustworthiness in FL should not be treated as a static property established at enrollment or training time, but as a continuously observable and verifiable state throughout the federation lifecycle.

The FAIR principles evaluate whether data are \textbf{Findable, Accessible,
Interoperable, and Reusable}~\cite{Wilkinson2016}. However, these dimensions are often considered separately in practice. A client may remain available while
training on stale or poorly traced data; a FAIR-compliant dataset may be consumed
without an auditable policy decision; and non-IID (\emph{non-independent and
identically distributed}) behavior may be mistaken for malicious behavior when
the federation lacks contextual evidence about how an update was produced.
Similarly, a data product may be discoverable and accessible while providing
insufficient runtime evidence about its quality, completeness, or operational
condition. Such evidence does not solve these problems by itself, but makes
their context inspectable and trust claims verifiable.

This challenge becomes more significant when industrial FL operates across
dataspaces, which support sovereign and governed data sharing across
organizations~\cite{Nagel2021DataSpaces,Otto2022DataSpaces,ge2025xckan}.
Data may be exchanged as governed products under access, usage, and even
pricing conditions~\cite{IDSA_RAM}, requiring assessment of data quality,
interoperability, provenance, and compliance before access or federated
training. However, the corresponding observability signals are heterogeneous
and may themselves be sensitive. We therefore introduce an optional
participant-side \textbf{TinyLLM for local signal interpretation, contextual
evidence derivation, and natural-language explanation}. Raw data and telemetry
remain local, while evidence verification and federation operation remain
independent of TinyLLM availability.

\begin{figure*}[t]
\centering
\includegraphics[width=\textwidth]{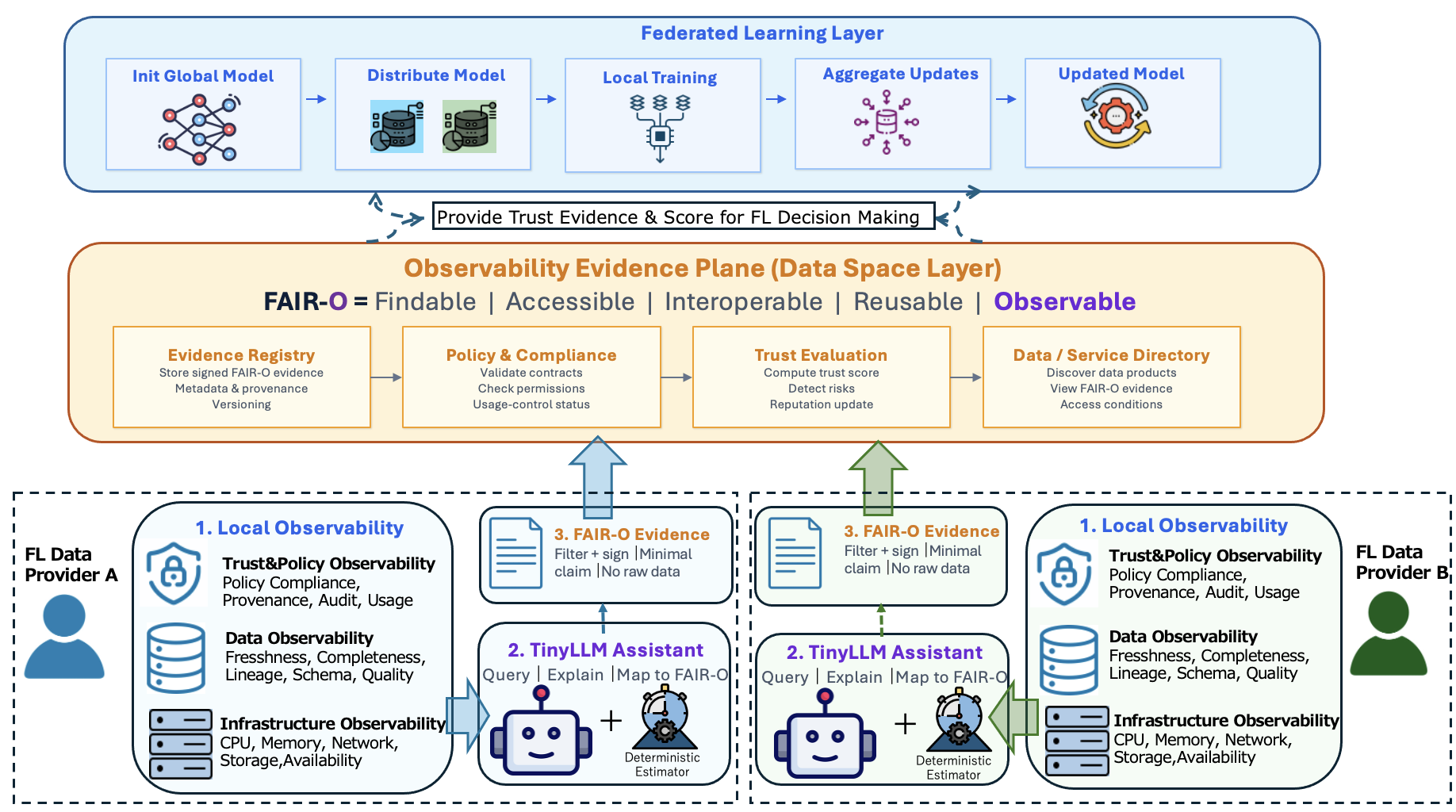}
   \caption{Overview of the \textbf{FAIR-O observability architecture} for trustworthy federated learning. Each FL data provider locally observes infrastructure, data, and trust-policy conditions. An optional participant-side TinyLLM supports local querying and explanation, while deterministic mechanisms generate minimal and signed FAIR-O evidence. The FAIR-O Evidence Plane manages this evidence at the Data Space layer and provides trust information to support federated-learning decisions without exposing raw data or operational telemetry.}
    \label{fig:fair-o-architecture}
\end{figure*}

We propose the FAIR principles with an explicit \textbf{Observability dimension}, which we term \textbf{FAIR-O} for Trustworthy Federated Learning. \textbf{FAIR-O considers not only whether a data asset satisfies the conditions for discovery, access, integration, and reuse, but also whether these conditions can be continuously assessed through verifiable operational evidence}. We structure this observability into three levels: \textbf{infrastructure observability}, covering computational resources, network conditions, availability, storage, and service health; \textbf{data observability}, covering freshness, completeness, accuracy, lineage, and schema drift; and \textbf{trust observability}, covering policy compliance, sovereignty monitoring, usage auditing, provenance verification, and derived trust indicators.

\subsection{FAIR-O Observability Architecture}

Figure~\ref{fig:fair-o-architecture} presents the proposed architecture.
The design separates local observability and evidence generation from the federated-learning workflow through an FAIR-O Evidence Plane.
The main principle is that raw industrial data and operational telemetry remain within each participant boundary, while only minimal and signed FAIR-O evidence
is exposed to the federation.

\subsubsection{Three Local Observability Levels}
Each FL data provider continuously observes its local operational environment along three complementary dimensions. 

\textbf{L1: Infrastructure observability} captures the operational condition of the
participant, including CPU and memory utilization, network status, storage
availability, service availability, and related infrastructure metrics.

\textbf{L2: Data observability} characterizes the current state and quality of the
local training data, including freshness, completeness, lineage, schema
consistency, and other data-quality indicators. These signals are particularly
important in industrial FL because a participant may remain technically
available while training on stale, incomplete, or structurally changed data.

\textbf{L3: Trust and policy observability} captures evidence related to policy
compliance, provenance, usage control, and audit events. This dimension extends
conventional infrastructure and data monitoring toward continuous assessment of whether a participant operates under the conditions expected by the
federation or by an associated Data Space contract.

\begin{figure*}[t]
\centering
\includegraphics[width=\linewidth]{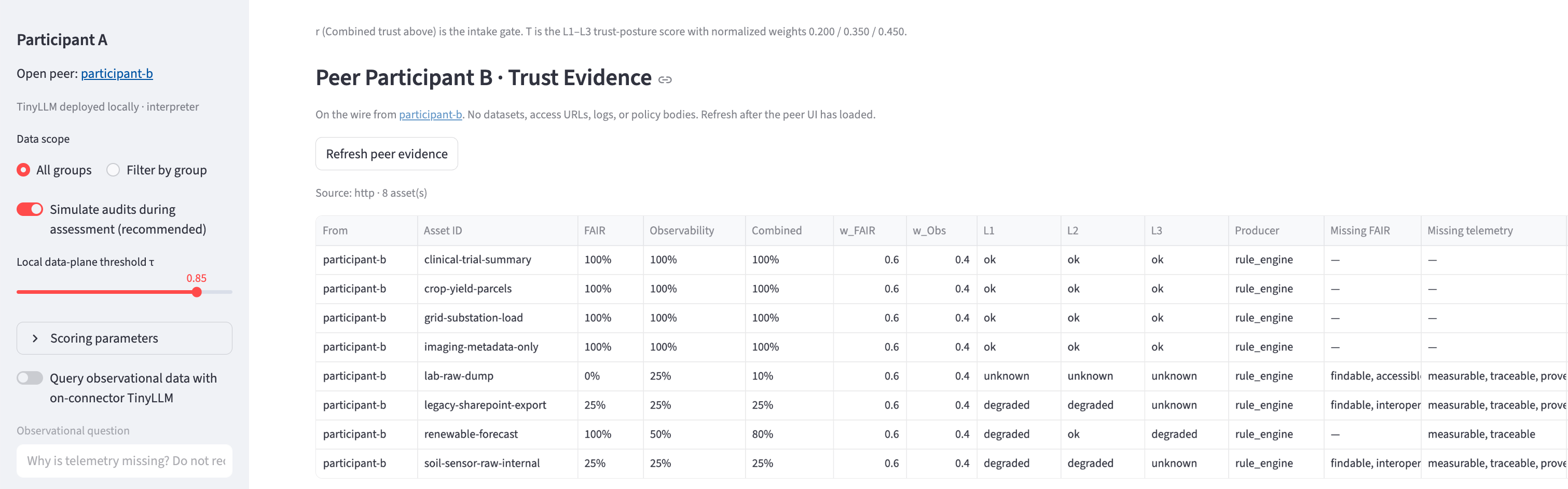}
   \caption{Participant A's view of Participant B's signed Trust Evidence}
    \label{fig:fair-overview}
\end{figure*}

% \begin{figure*}[t]
% \centering
% \begin{minipage}[t]{0.495\textwidth}
%   \centering
%   \includegraphics[width=\linewidth]{img/dashboard-air-quality.png}\\[-1mm]
%   \footnotesize (a) FAIR-complete but telemetry-poor air-quality profile.
% \end{minipage}\hfill
% \begin{minipage}[t]{0.495\textwidth}
%   \centering
%   \includegraphics[width=\linewidth]{img/dashboard-highway-flow.png}\\[-1mm]
%   \footnotesize (b) Fully instrumented highway-flow profile.
% \end{minipage}
% \caption{Participant-A dashboard case. The same transparent scoring policy
% distinguishes a telemetry blind spot from a fully instrumented asset. The
% pyramid exposes which observability level causes the difference; no dataset
% payload is displayed or included in peer Trust Evidence.}
% \label{fig:fair-dashboard}
% \end{figure*}

\subsubsection{Participant-Side TinyLLM Assistant}

A \textbf{TinyLLM assistant} can be deployed locally at individual
participants. This participant-side deployment is motivated by the privacy and data-sovereignty requirements of industrial FL. The goal is not to send private observability data to an LLM, but to bring a lightweight LLM to where the private observability data reside. 
The small computational footprint of TinyLLMs makes local deployment feasible, allowing observability signals to be processed within the participant's trusted environment without exposing raw data or telemetry to an external LLM service.

The locally deployed TinyLLM therefore provides a natural-language intelligence
interface over otherwise heterogeneous observability signals. Operators can
query system and data conditions, interpret anomalies, explain policy-related
events, and map observations to FAIR-O dimensions, while the underlying data
remain locally controlled. In this way, TinyLLM enables participants to derive
human-readable conclusions from private observability information without
requiring that information to be externally accessed or centralized.

Deterministic local components remain responsible for rule evaluation, measurement, filtering. Optionally, TinyLLM serves as a local reasoning and explanation layer over these mechanisms.
This separation ensures that FAIR-O evidence remains independently verifiable and available even when the TinyLLM component is unavailable or produces an unreliable explanation.

\subsubsection{Evidence Generation}

Local observability signals are transformed into FAIR-O evidence
before leaving the participant environment. The transformation follows the
principle of data minimization: raw logs, traces, industrial data, and detailed
telemetry are not exported.

\subsubsection{Evidence Plane}

The FAIR-O Evidence Plane operates at the Data Space layer to manage
and evaluate observability evidence across FL participants. The
\textbf{Evidence Registry} stores signed FAIR-O evidence with its metadata and
provenance. \textbf{Policy \& Compliance} validates evidence against contracts,
permissions, and usage-control requirements. \textbf{Trust Evaluation} derives
trust indicators and identifies potential risks or anomalies. Finally, the
\textbf{Data/Service Directory} links data products and services with their
access conditions and FAIR-O evidence. Together, these components provide
evidence-based trust information for data access and FL decision making without
exposing raw data or operational telemetry.

% The evidence plane therefore complements the conventional Data Space control
% and data planes. Rather than moving raw data to establish trust, it enables
% participants to exchange verifiable evidence \emph{about} the operational,
% data-quality, and governance conditions of those data.

\subsubsection{Integration with Federated Learning}
The federated-learning layer operates above the evidence plane using the standard workflow of global-model initialization, model distribution, local training, update aggregation, and global-model update. FAIR-O evidence provides additional context for FL decision making. For example, the federation may use evidence to support participant selection, round admission, update weighting, anomaly handling, or exclusion of a participant whose data or operational conditions no longer satisfy predefined
requirements. Conceptually, the resulting design follows a simple principle:
\begin{quote}
\emph{FAIR-O architecture enables observability of the training process and provide trust to support decisions}
\end{quote}

\begin{figure}[t]
\centering
\includegraphics[width=1\linewidth]{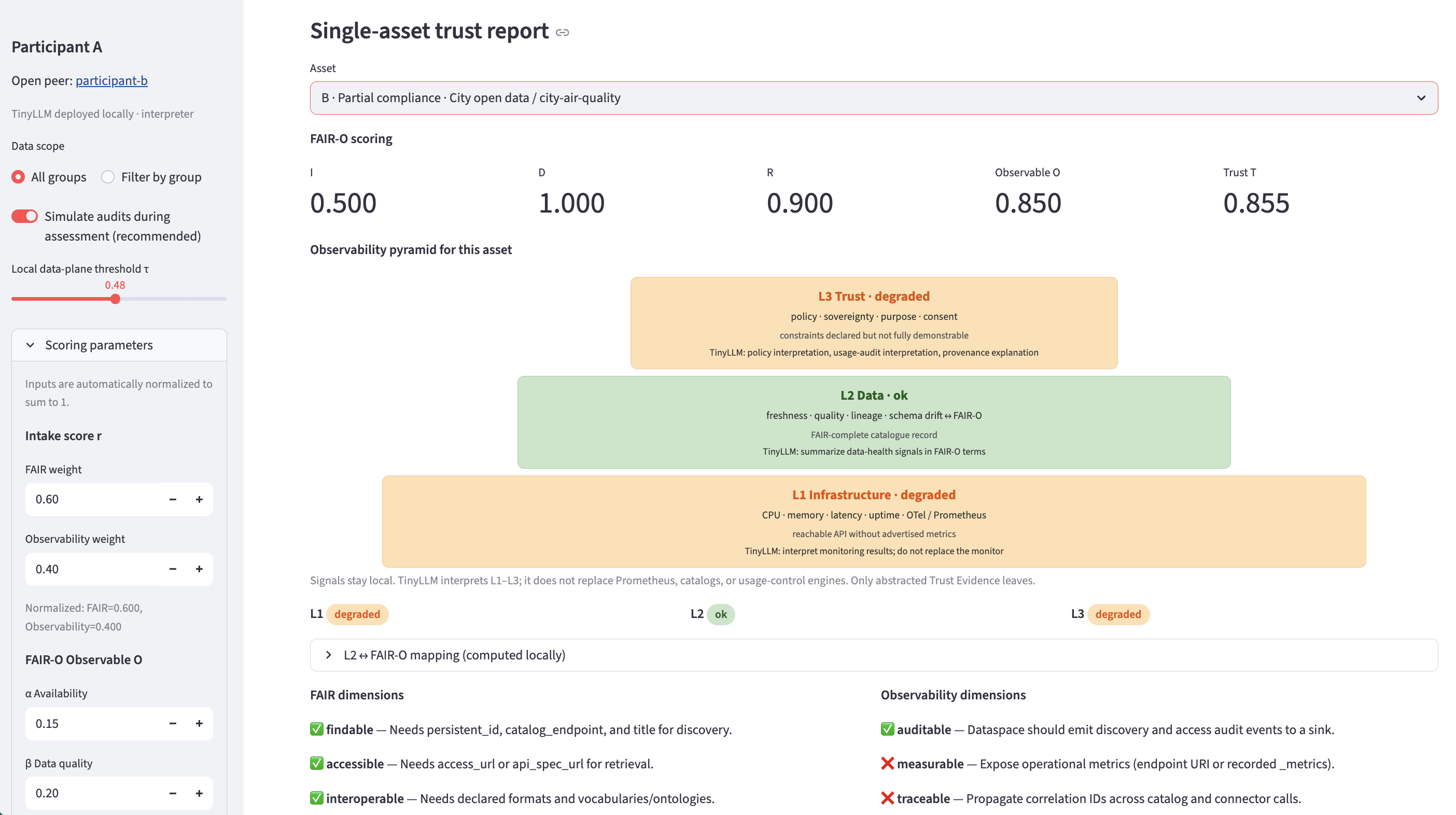}
   \caption{Participant-A dashboard case, showing the federation overview and the FAIR-O scores of data assets owned by Participant B. Participant A can inspect participant-level trust evidence and asset-level observability scores without revealing raw data.}
    \label{fig:fair-dashboard}
\end{figure}

\section{Prototype Grounding}

\label{sec:experiments}

We conduct two complementary experiments based on Python. The first evaluates whether FAIR-O
Trust Evidence can be generated, visualized, and exchanged without exposing
local data. The second evaluates executable federated learning when
participants have the same feature schema but different data quantity and
quality.

\subsection{Experiment 1: FAIR-O Evidence and Visualization}
Two logical participants : A and B can be launched as separate dashboard and evidence-API processes, with
distinct signing secrets and disjoint catalogues (six and eight synthetic
assets, respectively). Each dashboard constructs its own localhost TinyLLM
client. In the evaluated single-host configuration, both clients address the
same Ollama 0.11.2 service running the exact model tag
\texttt{qwen2.5:1.5b}~\cite{Qwen25}: a 1.5-billion-parameter Qwen2 model with a
32,768-token context window and GGUF \texttt{Q4\_K\_M} quantization. Thus, the
prototype demonstrates two logically separate clients, not two independently
hosted model servers. If Ollama is unavailable, deterministic summaries remain
available.

Figure~\ref{fig:fair-overview} shows the 
participant A's dashboard rendering the Trust Evidence pulled from participant B's
\texttt{/trust-evidence} endpoint. Each row of the overview table corresponds
to one asset that B publishes and carries only abstracted indicators: the
FAIR ratio, the observability ratio, and the combined trust score under the
declared weights ($w_{\mathrm{FAIR}}$, $w_{\mathrm{Obs}}$), the three
pyramid-level labels (L1 infrastructure, L2 data, L3 trust), the enumerated
missing FAIR fields and missing telemetry signals, a timestamp, and the HMAC
signature. Selecting an
asset expands the full exchanged payload; this is the entirety of what A can
learn about B's data before requesting access. The view therefore
operationalises the sovereignty constraint of the two-plane architecture:
A obtains enough signed, per-asset evidence to drive a local intake decision,
while B's raw records, access URLs, telemetry logs, and policy bodies never
leave B's premises.

% A deterministic rule engine emits summary Trust Evidence and six signed
% metric-level records per asset. The two APIs expose complementary views:
% \texttt{/trust-evidence} returns scores, missing dimensions, pyramid labels,
% timestamps, and signatures, whereas \texttt{/fair-o-evidence} returns metric
% values, thresholds, status labels, evidence hashes, and signatures. Outbound
% schemas contain no access URLs, payload fields, policy bodies, or raw logs;
% top-level banned-field checks guard summary publication and peer pull.
% TinyLLM responses are not copied into either evidence schema. These are
% data-minimization controls, not a formal non-interference or privacy proof.

Figure~\ref{fig:fair-dashboard} shows an FAIR-O assessment example of Participant A.
The \texttt{city-air-quality} data asset of Participant B is FAIR-complete but lacks an operational
metrics endpoint and correlation identifier. Its infrastructure and trust
levels are therefore degraded.
The dashboard further allows operators to interactively adjust the weights of
infrastructure, data, and trust observability evidence and immediately inspect
their impact on participant trust scores and FL decisions. This enables
application-specific and human-in-the-loop exploration of FAIR-O weighting
strategies.

\subsection{Experiment 2: Federated Learning Monitoring}
\begin{figure}[t]
  \centering
  \includegraphics[width=\linewidth,
    height=0.32\textheight,
    keepaspectratio]{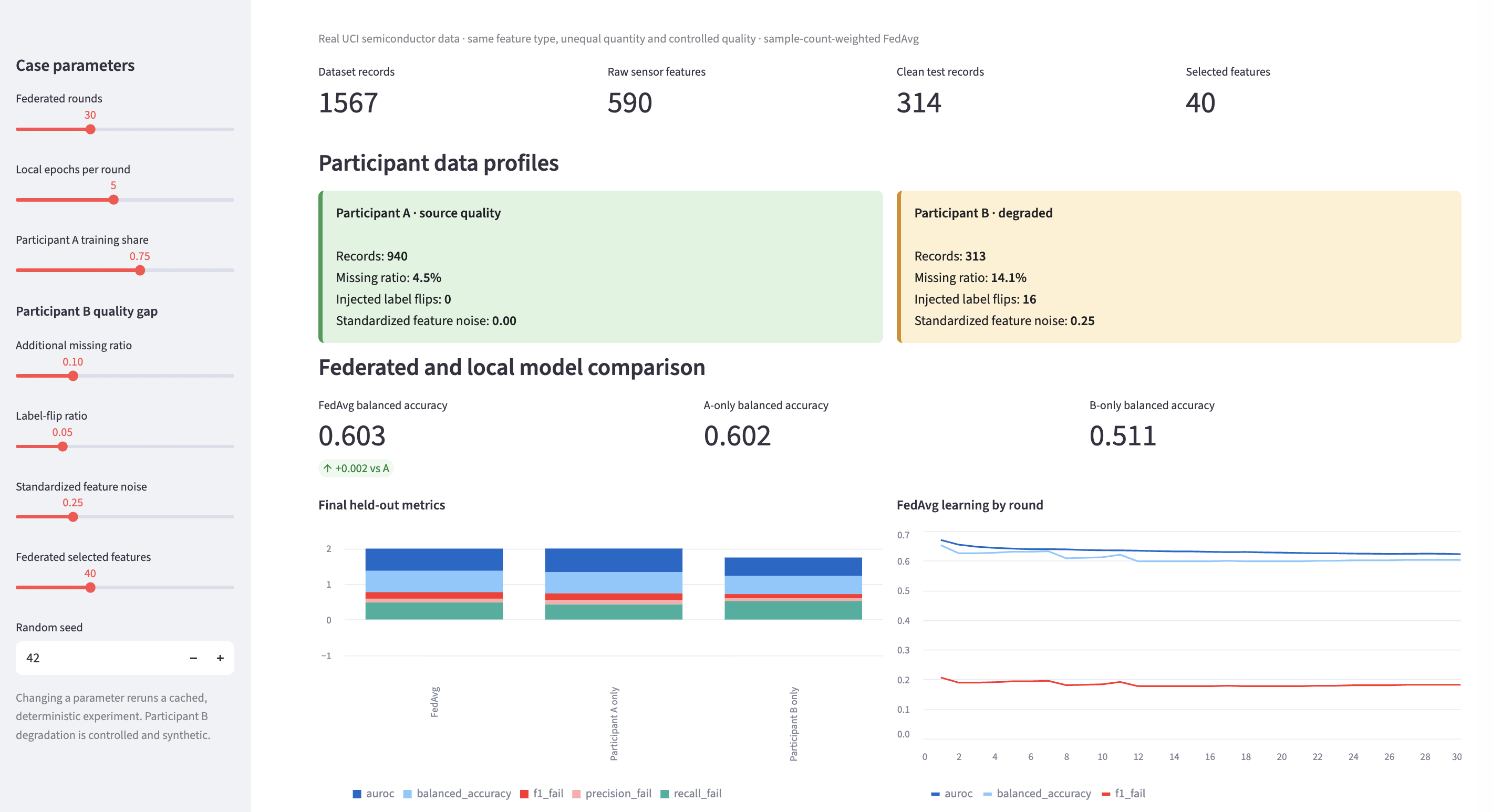}
  \caption{Dashboard output for the two-participant UCI SECOM experiment.
  Participant A contains 940 records with a 4.5\% source missing ratio.
  Participant B contains 313 records; controlled degradation increases its
  missing ratio to 14.1\%, flips 16 labels, and adds feature noise with
  standard deviation 0.25. The dashboard compares sample-count-weighted
  FedAvg with models trained only by A or B and shows the FedAvg learning
  trajectory over 30 rounds.}
  \label{fig:fedavg-dashboard}
\end{figure}
The second experiment uses the UCI SECOM semiconductor manufacturing
dataset~\cite{SECOM2008}. Figure~\ref{fig:fedavg-dashboard} illustrates the user interface to the FAIR-O Evidence Plane for output of a two-participant FL experiment. It jointly visualizes FL training progress, participant
observability conditions, signed evidence, policy-compliance status, trust
indicators, risks, and available data products. Operators can compare
participants, inspect anomalous conditions and their supporting evidence, trace
policy and evidence events, and understand the trust information used for FL
participation or aggregation decisions. Importantly, the dashboard exposes
derived FAIR-O evidence rather than raw participant data or telemetry, making
the distributed trust state of the federation observable, interpretable, and auditable.

The experiments validate different parts of the proposed architecture.
Experiment~1 demonstrates local FAIR-O assessment, explainable visualization,
evidence minimization, and peer verification. Experiment~2 demonstrates actual
FedAvg training over data with unequal participant quality. The two
paths are currently implemented separately: FAIR-O scores do not yet alter FL
participant automatic selection or aggregation weights. The next step is to bind signed FAIR-O evidence to each FL round and compare
ordinary FedAvg with evidence-aware admission and weighting.

\section{Conclusion}

We introduced \textbf{FAIR-O} as an observability extension of FAIR principle for
trustworthy federated learning. Participants retain raw data, logs, and
policies locally while sharing only minimized, signed \textbf{Trust Evidence}.
Evidence can be generated through deterministic rules, assisted by a local
TinyLLM for signal interpretation, contextual evidence derivation, and
natural-language explanation. We further implemented two experimental dashboards to demonstrate FAIR-O
observability and interactive trust assessment. Future work will incorporate
Trust Evidence into participant selection and aggregation, explore adaptive
weighting strategies, and evaluate their impact on accuracy, robustness,
fairness, privacy, and overhead. Ultimately, FAIR-O shifts FL trust from a
static property toward a \textbf{continuously observable and verifiable state}.

% Future work will investigate adaptive weighting strategies for FAIR-O evidence, where infrastructure, data, and trust observability signals are assigned different importance according to the FL task, participant context, and training round. The resulting evidence weights can then be incorporated into participant selection, update weighting, or robust aggregation.

% Client updates will be assessed using sample size with FAIR-O evidence. The present experiment uses ordinary sample-count-weighted FedAvg; evidence-aware admission and weighting remain future work. 

\bibliographystyle{IEEEtran}
\bibliography{refs}

\end{document}